\documentclass[final,1p,times,twocolumn]{elsarticle}

\usepackage{geometry}  
\usepackage{verbatim} 
\usepackage{graphicx}											
\usepackage{amssymb}
\usepackage{caption}
\usepackage{subfigure}
\usepackage{amsmath}

\usepackage{delarray} 
\usepackage{amsthm}

\usepackage[noend]{algpseudocode}
\usepackage{algorithmicx,algorithm}

\usepackage{stfloats}
\usepackage{booktabs}
\usepackage{graphicx}

\newtheorem{remark}{Remark}

\author[Address1]{Qiming\ Zou\corref{cor1} }
\ead{qimingzou@aliyun.com}

\author[Address1]{Ling\ Wang\corref{cor2}}
\ead{wangling@hit.edu.cn}

\author[Address2]{Ke\ Lu\corref{cor3}}
\ead{luke.airoot@gmail.com}

\author[Address2]{Yu\ Li\corref{cor3}}
\ead{anna.l@aliyun.com}

\cortext[cor1]{Corresponding author}
\cortext[cor2]{Principal corresponding author. 
Department of Computer Science and Technology, Harbin Institute of Technology, No.92, Xidazhi Street, Nangang District, Harbin City, Heilongjiang Province, China.\\
Email address: qimingzou@aliyun.com (Qiming Zou), wangling@hit.edu.cn (Ling Wang).
}

\address[Address1]{Department of Computer Science and Technology, Harbin Institute of Technology, China}	
\address[Address2]{Department of Management Science and Engineering, Anhui University of Technology, China}

\makeatletter
\def\ps@pprintTitle{%
   \let\@oddhead\@empty
   \let\@evenhead\@empty
   \let\@oddfoot\@empty
   \let\@evenfoot\@oddfoot
}
\makeatother

\begin{document}

\begin{frontmatter}

\title{Sample-Efficient Policy Learning based on Completely Behavior Cloning } 

\begin{abstract}
Direct policy search is one of the most important algorithm of reinforcement learning.
However, learning from scratch needs a large amount of experience data and can be easily prone to  poor local optima.
In addition to that, a partially trained policy tends to perform dangerous action to agent and environment.
In order to overcome these challenges, this paper proposed a policy initialization algorithm called Policy Learning based on Completely Behavior Cloning (PLCBC).
PLCBC  first transforms the Model Predictive Control (MPC) controller into a piecewise affine (PWA) function using multi-parametric programming, and uses a neural network to express this function.
By this way, PLCBC can completely clone the MPC controller without any performance loss, and is totally training-free.
The experiments show that  this initialization strategy can help agent learn at the high reward state region, and converge faster and better.  

\end{abstract}
\end{frontmatter}

\section{Introduction}
Deep reinforcement learning  is becoming increasingly popular for tackling challenging sequential decision making problems, and has been shown to be successful in solving a range of difficult problems, such as games \cite{Silver:2016aa, Silver:2017aa}, robotic control \cite{levine2016end} and locomotion \cite{schulman2017proximal, heess2017emergence}.
One particular appealing prospect is to use deep neural network parametrization to minimize the burden for manual policy engineering \cite{gu2016q}.

Deep neural network is a general and flexible representation of policy, which can represent complex behaviors \cite{levine2013guided}.
However, using deep neural networks to perform policy search from scratch is exceedingly challenging for two reasons \cite{levine2016end}. 
First,  learning such complex, nonlinear policy with standard policy gradient methods can require a huge number of iterations, and be disastrously prone to poor local optima.
The second obstacle to use Deep Reinforcement Learning (DeepRL) in the real world is that, although a fully trained neural network controller can be very robust and reliable, a partially trained policy can perform unreasonable and even unsafe actions. 
This can be a major problem when the agent is a mobile robot or autonomous vehicle and unsafe actions can cause damage to the robot or its surroundings.

We address these challenges by developing an algorithm called Policy Learning based on Completely Behavior Cloning (PLCBC), a method for training complex polices by initializing the weights of deep neural network with a trajectory-optimization computational teacher, namely, model predictive control (MPC) \cite{XI2013}.
To be more specific, firstly the MPC controller (an implicit control law) is transformed into an explicit piecewise affine (PWA) function using multi-parametric programming technique.
Based on this PWA function, the MPC controller can be easily incorporated into a deep neural network.
MPC is a major method for  optimal control of dynamic system, which performs really well in a broad range of sequential decision problems. Therefore, it is expected to improve the efficiency of the DeepRL by exploring high-reward regions.

This method has several appealing properties.
First,  stabilizing MPC controllers is easier than that of arbitrary policies.
Since the policy is initialized with MPC controller,  this mechanism can be a notable safety benefit when the initial parameterized policy is unstable, and make the policy stay away from the poor local optima.
Second, since our algorithm can completely clone the behavior of MPC under entire state space,  there is no state distribution inconsistent problem (or compound error)  that is common issue of imitation learning algorithm \cite{Schroecker2017, Wang2017}.
Last, the initial weight of our network is coded by the PWA function, therefore in the phase of imitation, PLCBC is totally training free that means additional sample complexity is removed.

The rest of this paper is organized as follows.
In Section 2, a piecewise affine function of the MPC controller is obtained using multi-parametric technique.
The main procedure of transforming the PWA function into deep neural network is developed in Section 3. 
In Section 4, we will introduce a modified reinforcement learning algorithm that is adapted to PLCBC.
In Section 5, three control problems are given to illustrate the effectiveness of the proposed controller.
Finally, the conclusion is drawn and our future work will be declared in Section 6.

\section{Related Work}
Broadly speaking, deep reinforcement learning can be roughly classified into two categories. First, reward based methods, including deep Q learning \cite{Mnih2013}, policy gradient algorithm \cite{DDPG}. Second, imitation based methods, including  naive supervised learning, Dataset Aggregation (DAgger) \cite{Dagger2011}, Guided Policy Search (GPS) \cite{Levine}.

Reward based methods update the parameters to maximize the accumulate reward.
Usually, by using a number of training episodes, the explicit knowledge of the underlying model is unnecessary to iteratively improve the policy from real-world experience \cite{fazel2018global}.
However, this kind of approaches is quite challenging when the reward function is hard to design or the reward signal is sparse \cite{Florensa2017}.
Besides, reward based methods can be really sample inefficient.
They all tend to require very large number of samples to learn the high-dimensional neural network policy from scratch, which is the bottleneck of their applicability  to real world \cite{gu2016q}.
Furthermore, these methods can be easily prone to local optima, making it very difficult to  find a good solution \cite{levine2013guided, Levine}.

In the imitation algorithm, the learner tries to mimic an expert's action in order to achieve the best performance \cite{ross2010efficient}.
Typically, these methods do not need to design a reward function, and are dramatically more sample efficient.
Nevertheless, a viable human or computational expert is required to generate labeled samples \cite{Kahn}.
Since imitation learning needs to query the expert frequently, it will require extremely large amount of expert demonstrations to learn, especially in the continuous control scenarios.
In addition, demonstrations from human expert can be quite expensive. Therefore the mainly focus of this work is the algorithms whose supervision comes from a computational expert.
One simple way to imitate computational expert is to use the slow planning based controller like Monte-Carlo tree search planning to provide training data for a neural network, and the agent directly learns the expert's action in a naive supervised learning fashion \cite{Argall2009, guo2014}.
Although this kind of methods are appealingly simple, they can lead to a problem called compounding error, which means that even a small error can be accumulated to a disastrous consequence, leading the agent away from the region of the state space where it was given examples, leading to unrecoverable failures \cite{Kahn}.
Ross and Bagnell showed the number of errors made by the agent trained with naive supervised learning, in the worst case, can scale quadratically with the time horizon of the task \cite{Bagnell2010}.
Besides, naive supervised learning can be really sensitive to the randomness of real-world system and the model error of computational expert. 
In the case of  DAgger \cite{Dagger2011}, the learner mimics the control action by iteratively gathering more examples from the supervisor in states the robot encounters. 
At each iteration, the agent trains a policy based on the existing examples, then rolls out that policy.
The supervisor provides demonstrations for all states the agent visits, and the agent combine them with the old examples for the next iteration.
In DAgger, under certain conditions, the number of errors scales only linearly with the time horizon of the task.

However, DAgger needs to perform the partial learned policy in real-world, which will be dangerous to the environment or the agent, especially in the safe-critical scenarios.
Beyond that, under the DAgger framework, the performance of the fully trained policy will not be better than computational expert.
Guided Policy Search uses differential dynamic programming (DDP) to generate “guiding samples” to assist the policy search. 
The parameterized policy never needs to be executed on the real system, because interactions between agents and environment during training are done using time-varying linear-Gaussian controllers \cite{Levine}.
While prior applications of guided policy search rely on a learned model of system dynamic,  it can cause extremely error if the learned model is inaccurate \cite{Zhang, levine2014b, levine2015b}.
In order to overcome this challenge, Zhang uses MPC controller as the computational expert, and takes fully advantage of its replanning framework to reduce the error caused by model error \cite{Zhang}.

Aforementioned imitation based algorithms all need to sample the experience data from the interaction between the computational expert or real-world system, which can be really time-consuming and inefficient.
In this paper, we  propose a totally offline behavior cloning algorithm, namely PLCBC.
This method directly transforms the MPC controller into a deep neural network, and then the policy gradient based reinforcement learning algorithm will be implemented upon it.
The initialized policy will not execute dangerous action to environment or agent itself, and can effectively avoid the terrible local optima.

\section{PWA Function Transformed From MPC}
In this section, the corresponding PWA function of MPC is obtained using multi-parametric technique.
This resulting explicit function  is usually referred as Explicit Model Predictive Control (EMPC) controller \cite{Bemporad2002}.

Consider a discrete linear time-invariant (LTI) system with box constraints:
\begin{equation}\label{eq:12}
\begin{split}
& x_{t+1}=Ax_t+Bu_t \\
& y_t=Cx_t \\
& x_t\in \textbf{X}:=\{x\in \mathbb{R}^n:x_{min}\leq x\leq x_{max}\}\\
& u_t\in \textbf{U}:=\{u\in \mathbb{R}^m:u_{min}\leq u\leq u_{max}\}
\end{split}
\end{equation}
where $A\in \mathbb{R}^{n\times n}$, $B\in \mathbb{R}^{n\times m}$ and $C\in \mathbb{R}^{p\times n}$.

The typical MPC optimization problem with time horizon $N$ can be written in the following compact matrix form:  
\begin{equation}
\label{eq:objfunc}
\begin{split}
&J^*(U, x_t)=\frac{1}{2}x_t^TYx_t+\min_{U}\left\{\frac{1}{2}U^TQU+x_t^TRU\right\}\\
&s.t.\quad GU\leq W+Ex_t
\end{split}
\end{equation}
where $U=[u_t^T,u_{t+1}^T,\cdots,u_{t+N-1}^T]^T$ is optimization vector and $Y, Q, R, G, W, E$ are constant matrices of appropriate dimension.

The concept of multi-parametric  programming consists in considering the state vector $x_t$ as a parameter and determining the optimal input vector $u^*_t$ as an explicit function of the state vector $x_t$.
Considering problem (\ref{eq:objfunc}), optimal input sequence $U^*$ is continuous and piecewise affine on polyhedra $P^r$, $i.e.$ if $x_t \in P^r$, $r=1, \cdots, N_p$, then
\begin{equation}
\label{eq:pwa}
\begin{split}
&U^*(x_t) = F^rx_t+g^r\\
&u^*_t =  (I, 0, \cdots, 0)U^*(x_t)
\end{split}
\end{equation}
where $N_p$ is the number of linear pieces, and the polyhedra $P^r$ is described by
\begin{equation}
\label{eq:region}
P^r = \{x_t|H^rx_t\leq k^r\}
\end{equation}
where $H^r\in \mathbb{R}^{n_c^r\times n}$ and $k^r\in \mathbb{R}^{n_c^r}$, $n_c^r$ is the number of linear constraints corresponding to state region $P^r$.
Note that state space has been divided into a polyhedral region set $P=\{P^1,P^2,\cdots,P^{N_p}\}$, $\bigcup_{r=1}^{N_p}P^r = \textbf{X}$.

Therefore, online computation of MPC can be simplified to a point location problem.
Given a query point $x_t\in \mathbb{R}^{n}$, the problem is to find  an appropriate integer number $i(x_t)\in \{0, 1, \cdots, N_p\}$ such that $i(x_t) = 0$ if $x\notin P$ and $1\leq i(x_t) \leq N_p$ if $x\in P^{i(x_t)}$.
\begin{remark}
In this work, it is assumed that during training, the true underlying states $x_t$ have been accessed.
This additional assumption allows us to use EMPC to generate optimal actions.
\end{remark}
\begin{remark}
If the dynamic of the system is nonlinear or unknown, we still can divide the action and state space into several subspace, and use a piecewise linear equation to appropriate the real system \cite{1184260}.
\end{remark}

\section{Neural Network Transformed From PWA Function}

\subsection{Main Idea}
According to the equation (\ref{eq:pwa}),  it is obvious that if one can get the number $i(x_t)$, the optimal control law can be easily calculated.
Inspired by this idea, PLCBC algorithm is proposed to clone MPC controller without loss of optimality.
To be more specific, PLCBC consists of two networks, location network and policy network.
Location network is used to express the structure of polyhedral region set with a neural network and achieve the process of point location.
It takes state $x_t$ as input and outputs the one-hot encoding of $i(x_t)$.
Policy network is used to express the linear control law of each state region which maps specific state to deterministic action.
In the next section, it is presented that how to assign the weights and bias of both subnetworks to express the MPC controller.
\subsection{Neural network architecture}
\subsubsection{Location Network}
\begin{figure}[htp]
\begin{center}
\includegraphics[width=0.7\textwidth]{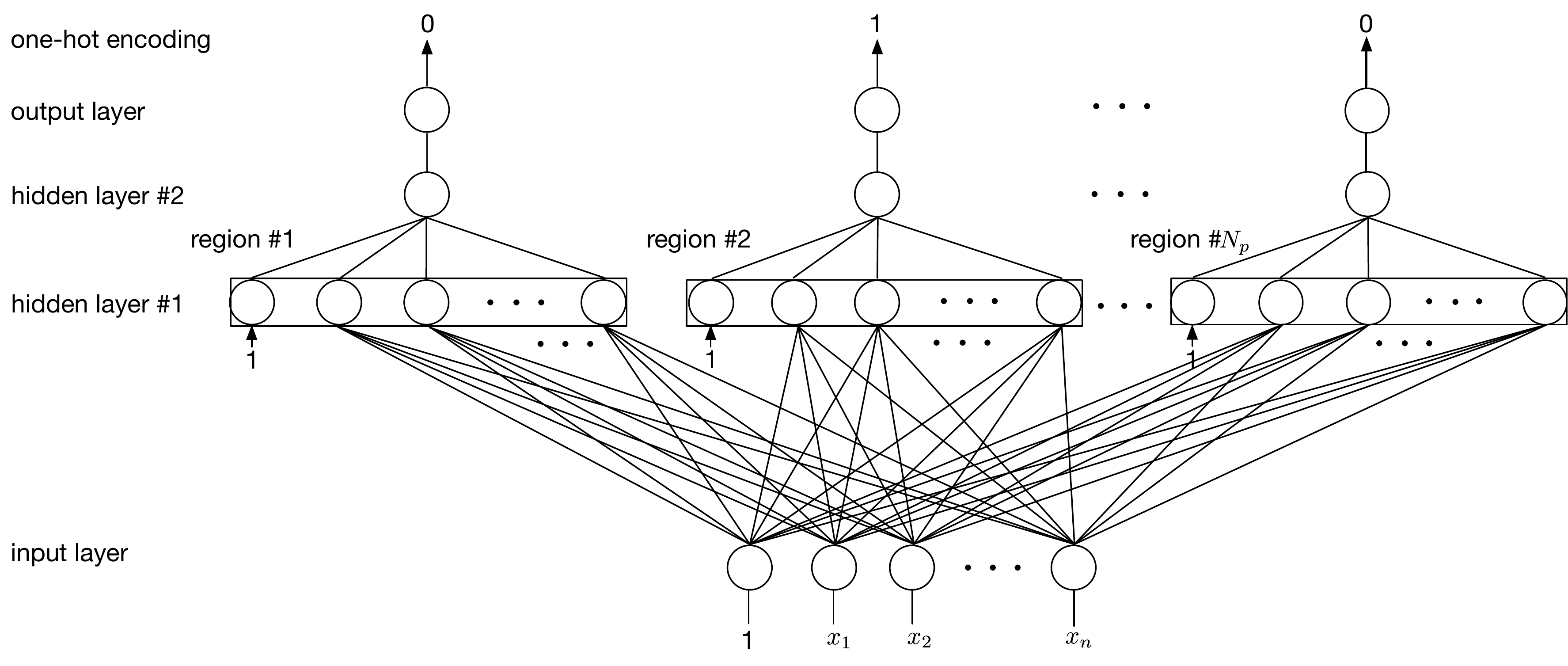}
\caption{PLCBC architecture.}
\label{PLCBC}
\end{center}
\end{figure}

In most cases, training a neural network to map input to one-hot encoding of  label (index of state region in this case) will use supervised learning based methods.
However, since each state region can be represented by a set of linear inequalities, we will directly achieve the location function by coding the weights of location network. 
The network architecture is composed of 4 layers named input layer, $1^{th}$ hidden layer, $2^{th}$ hidden layer and output layer shown in Fig.~\ref{PLCBC}. 

In the input layer except for $n$ input variables a node representing external bias 1 is added.
The first and second hidden layers are created to perform as location  function whose weights and structure are determined by the PWA function transformed from MPC controller.
In the output layer, there are $N_p$ neurons to output the one-hot encoding of region number.
The location network is composed of a specific type of neurons using the activation function $\sigma^{loc}$:
\begin{equation}
\label{aa}
\sigma^{loc}(z)=\left\{
\begin{aligned}
1 & , & z \ge 0  \\
0 & , & z < 0 
\end{aligned}
\right.
\end{equation}
However discrete activation function will make it hard for neural network to be updated via gradient-based methods.
Therefore, $\sigma^{loc}$ can be replaced by a simple sigmoid function during training, we still can output the one-hot encoding of region index with a softmax layer. 

$W^{loc}_ {i, i + 1}$ and $B^{loc}_i$ represent the weight and the bias of the $i^{th}$ hidden layer respectively, then 
\begin{equation}
W^{loc}_{1,2}=
\begin{bmatrix}
-H^1, 
-H^2, 
\cdots, 
-H^{N_p}
\end{bmatrix}
^T
\end{equation}
\begin{equation}
B^{loc}_1=\begin{bmatrix}k^1, k^2, \cdots, k^{N_p}\end{bmatrix}^T
\end{equation}
\begin{equation}
W^{loc}_{2,3}=
\begin{bmatrix}
I_{n_c^1\times 1} & 0 & \cdots & 0\\
 0 & I_{n_c^2\times 1} & \cdots & 0\\
\vdots & \vdots & \cdots & \vdots\\
 0&  0 & \cdots&  I_{n_c^{N_p}\times 1}\\
\end{bmatrix}
\end{equation}
\begin{equation}
B^{loc}_2=\begin{bmatrix}-n_c^1, -n_c^2, \cdots, -n_c^{N_p}\end{bmatrix}^T
\end{equation}
where $I_{n_c^i\times 1}$ is a vector made of all ones, and $H$, $K$ come from equation (\ref{eq:pwa}).
Let $net^{loc}_{i}$, $out^{loc}_{i}$ be the input and output of $i^{th}$ hidden layer.
Based on the feedforward computation of neural network, $net^{loc}_{i}$ and $out^{loc}_{i}$ can be calculated as follows.

\noindent\textbf{Hidden layer $\#$1}
\begin{equation}
\label{eq:net^{loc}_1}
\begin{split}
net^{loc}_{1} & = W^{loc}_{1,2} x+B^{loc}_1\\
& = \begin{bmatrix}-H^1x+k^1, \cdots, -H^{N_p}x+k^{N_p}\end{bmatrix}^T\\
\end{split}
\end{equation}
Equation (\ref{eq:net^{loc}_1}) shows that $net^{loc}_{1}$ can be partitioned into $N_p$ blocks and each of them is a $n_c^{i} \times 1$ vector.

\begin{equation}
\begin{split}
out^{loc}_{1} & =\sigma^{loc}(net^{loc}_{1})\\
& = \begin{bmatrix}\sigma^{loc}(-H^1x+k^1), \cdots, \sigma^{loc}(-H^{N_p}x+k^{N_p})\end{bmatrix}
\end{split}
\end{equation}
Since
\begin{equation}
-H^{i(x_t)}x+K^{i(x_t)} \ge \mathbf{0}_{n_c^{i(x_t)}\times 1}
\end{equation}
and $j^{th}$ block $-H^jx+k^j,j \neq i(x_t)$ can not satisfy this inequality,
based on the definition of $\sigma^{loc}$,
\begin{equation}
\label{eq:i_x}
\sigma^{loc}(-H^{i(x_t)}x+k^{i(x_t)}) = I_{n_c^{i(x_t)}\times 1}
\end{equation}
while $\sigma^{loc}(-H^jx+k^j),j \neq i(x_t)$ is a $n_c^{i(x_t)}\times 1$ vector having some 0 bits.

\noindent\textbf{Hidden layer $\#$2}

For brevity, we only describe the $i^{th}$ block of $2^{th}$ hidden layer that corresponds to the $i^{th}$ output of the neural network. 
\begin{equation}
\label{eq:net^{loc}_2}
\begin{split}
net^{loc}_{2}(i) &= W^{loc}_{2,3}(i)out^{loc}_{1}+B^{loc}_2(i)\\
&=\begin{bmatrix}\mathbf{0},\cdots,I_{1\times k^i},\cdots,\mathbf{0}\end{bmatrix}
\begin{bmatrix}\vdots\\\sigma^{loc}(-H^ix+k^i)\\\vdots\end{bmatrix}
+\begin{bmatrix}\vdots\\ -k^i\\ \vdots \end{bmatrix}
\end{split}
\end{equation}

\begin{equation}
\label{eq:out^{loc}_h2}
out^{loc}_{2}(i)=\sigma^{loc}(net^{loc}_{2}(i))
\end{equation}

According to equations (\ref{eq:i_x})(\ref{eq:net^{loc}_2})(\ref{eq:out^{loc}_h2}), 
\begin{equation}
out^{loc}_{2}(i) =
\begin{cases}
1,& i=i(x_t)\\
0,& i\neq i(x_t)
\end{cases} 
\end{equation}

Therefore, based on the network introduced above,  the state region of particular input $x_t$ can be located by a simple feedforward computation.
Then, we need to use $i(x_t)$ to activate the particular policy subnetwork to output the action $u_t$.
\subsubsection{Policy Network}

\begin{figure}[htp]
\begin{center}
	\includegraphics[width=0.5\textwidth]{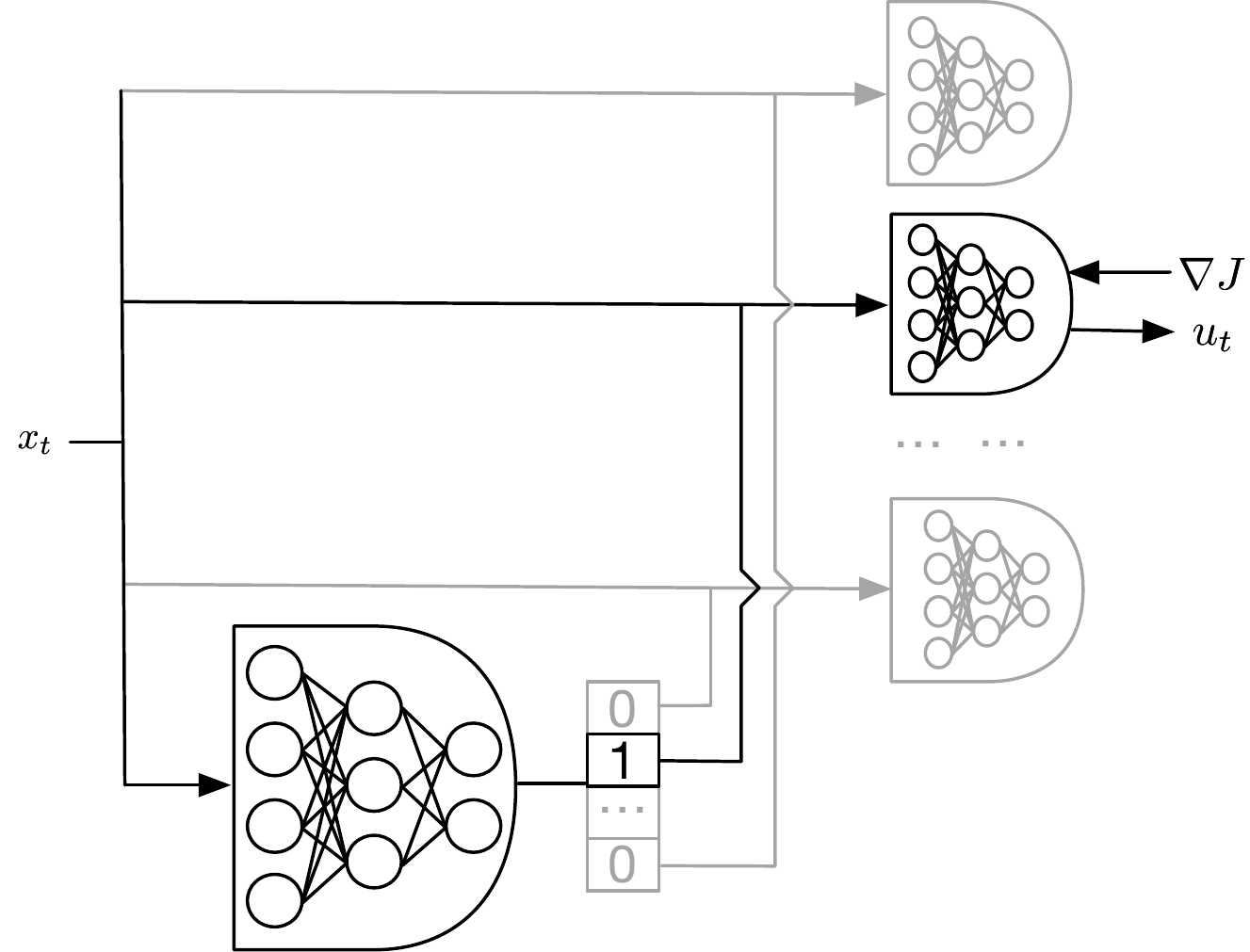}
	\caption{Block-diagram of PLCBC: given a state $x_t$, the location network outputs $i(x_t)$ which activate the corresponding policy subnetwork. State $x_t$ is propagated through this policy network, and the action $u_t$ is outputted. In the backward phase, the gradient message $\nabla J$ is back propagated through the active policy network to update the weights. }\label{fig:PLCBC}
\end{center}
\end{figure}

According to (\ref{eq:pwa}), we can directly use $N_p$ linear perceptrons $\pi^i(x_t|\theta^i),\ i=1, \cdots, N_p$ to express the optimal control law of MPC.
The weights and biases of  $i^{th}$ policy networks denoted as $W_{1,2}^{plc^i}$ and $B_{1}^{plc^i}$ can be assigned as follows:
\begin{equation}
W^{plc^i}_{1,2} = F^i
\end{equation}
\begin{equation}
B^{plc^i}_{1} = g^i
\end{equation}
where $F^i$ and $g^i$ come from equation (\ref{eq:pwa}).
Note that the activation function of these perceptrons need to be constant function $\sigma^{plc}(z) = 1 $, and 
\begin{equation}
out^{plc^i}_{1} = net^{plc^i}_{1} = W^{plc^i}_{1,2}x+B^{plc^i}_{1}
\end{equation}

Therefore in order to clone the PWA function, we consider to combine the location network and the policy network.
In such a network, we use a $N_p$-way junction to make a decision that which policy subnetwork the input vector will propagate through. 
The input of this routing junction is the one-hot encoding of region index $i(x_t)$ computed by the location network introduced before.
According to $i(x_t)$, the junction will lead the sate vector $x_t$ into the $i^{th}$ policy network if and  only if  $i(x_t)=i$.
Our multi-path architecture is illustrated in Fig.~\ref{fig:PLCBC}.
\begin{remark}
The total network size ($i.e.$, the number of total parameters) is proportional to $N_p$. 
If the underlying dynamic is highly non-linear, PLCBC may use much more parameters than the standard deep reinforcement learning algorithms.
One possible solution is to use the incremental multi-parametric algorithm which only generates linear policies  of encountered states \cite{GP2007}.
\end{remark}
\begin{remark}
PLCBC may perform poorly if the underlying dynamics is too complex so that highly non-linear policy network is required. This problem can be tackled by slightly changing the architecture to $out^{plc^i}_{1} = net^{plc^i}_{1}+g^i(x)$, where $net^{plc^i}_{1}$ is the (MPC-derived) linear policy term, and $g^i(x)$ is the non-linear residual part that is initialized to zero function and would be fine-tuned.
\end{remark}

\section{Policy Searching}
In the standard reinforcement learning setting, we take the combined neural network introduced above as actor network $\pi(x_t|\theta_\pi)$ that interacts with an environment $E$ over a number of discrete time steps. 
 At each time step $t$, the actor receives a state $x_t$ and outputs action $u_t$.
 In return, the environment generates the next state $x_{t+1}$ and a scalar reward $r_t$. 
$R_t = \sum_{k=0}^{T-t}\gamma^kr_{t+k}$   is the total accumulated return from time step $t$ to the end of this episode with discount factor  $\gamma\in(0,1]$. 
The goal of the learning is to maximize the expected accumulated return from each time step. 

Our focus is reinforcement learning in environments with continuous state and action space. 
Deep Deterministic Policy Gradient (DDPG) provides an attractive paradigm for continuous control \cite{DDPG}. 
Besides the actor network, DDPG also maintains a critic function $Q(x_t,u_t)$ that outputs the estimated Q-value of the current state $x_t$ and of the action $u_t$ given by the actor. 
As proposed in the DQN training, DDPG uses slowly changing target network $\bar{Q}$ and $\bar{\pi}$ to compute the Q-learning loss $L_Q$:
\begin{equation}
L _Q= \mathbb{E}_{x_t}(r_t+\gamma \bar{Q}(x_{t+1},\bar{\pi}(x_{t+1}|\theta_{\bar{\pi}})|\theta_{\bar{Q}})-Q(x_t,u_t|\theta_Q))^2
\end{equation}
The actor is updated by minimizing the loss: :
\begin{equation}
L _{\pi}=  \mathbb{E}_{x_t}[Q(x_t,u_t|\theta_Q)]+\lambda KL[\pi_{old}|\pi]
\end{equation}
In this loss function, the only difference from a standard DDPG algorithm is the inclusion of the KL-divergence term. 
This term aims to limit the change of current policy from old policy \cite{heess2017emergence, schulman2017proximal}.
In order to take advantage of MPC controller, the weight $\lambda$ of KL term is relatively large.

One challenge to use neural networks for reinforcement learning is that most optimization algorithms assume that the samples are independently and identically distributed. 
Obviously, when the samples are generated sequentially in an environment this assumption no longer holds. 
In DDPG, a replay buffer is used to address these issues. 
The replay buffer is a finite sized cache $R$. 
Transitions were sampled from the environment according to the exploration policy and the tuple $ (x_t, u_t, r_t, x_{t+1}) $  was stored in the replay buffer.  
At each time step the actor and critic are updated by sampling a mini-batch uniformly from the buffer \cite{DDPG}. 

\begin{algorithm}[t]
\caption{Reinforcement Learning for multiple Policy Subnetwork} 
\begin{algorithmic}[1]
\State{Initialize location network and all the policy subnetworks $\pi^i(x|\theta_{\bar{\pi}^i})$, $i = 1,\ \cdots,\ N_p$ with PWA function obtained from MPC controller}
\State{Randomly initialize the shared critic network $Q(x,u|\theta_Q)$}
\State{Initialize target network  $\bar{Q}$ and $\bar{\pi}^i$ with weights $\theta_{\bar{Q}}\leftarrow \theta_Q$}, $\theta_{\bar{\pi}^i}\leftarrow \theta_{\pi^i}$, $i = 1,\ \cdots,\ N_p$ 
\State{Randomly initialize replay buffer $R^i$, $i=1,\ \cdots,\ N_p$}
\For{episode=1,$\cdots$, M} 
　　\State Receive initial observation state $x_1$
	\For{t=1, $\cdots$, T} 
		\State Locate the region of the query state $x_t$ and activate the corresponding policy $\pi^i(x|\theta^i)$.
　　		\State Select action $u_t$ according to the current policy and exploration noise 
		\State Execute action $u_t$; observe reward $r_t$ and new state $x_{t+1}$
		\State Store transition $(x_t, u_t, r_t, x_{t+1})$ in $R^i$ belonging to the current policy 
 
	\State Update critic and actors using DDPGLearn($Q$, $\bar{Q}$, $\pi$, $\bar{\pi}$, $R$, $N_p$)
	\State Update the target networks:
			\begin{equation*}
			\begin{split}
				&\theta_{\bar{Q}}\leftarrow \tau \theta_Q+(1-\tau)\theta_{\bar{Q}}\\
				&\theta_{\bar{\pi}^i}\leftarrow \tau \theta_{\pi^i}+(1-\tau)\theta_{\bar{\pi}^i},\ i = 1,\ \cdots,\ N_p 
			\end{split}
		\end{equation*}	
			\EndFor 
\EndFor
\end{algorithmic}
\end{algorithm}

\begin{algorithm}[t]
\caption{DDPGLearn} 
\hspace*{0.02in} {\bf Input:} 
critic network $Q(x_t,u_t)$, target critic network $\bar{Q}(x_t,u_t)$, policy subnetworks $\pi^i(x_t|\theta_{\pi^i})$, \\target policy subnetworks $\bar{\pi}^{i}(x_t|\theta_{\bar{\pi}^i})$, replay buffer $R^i$, and policy number $N_p$, $i = 1,\ \cdots,\ N_p$\\
\hspace*{0.02in} {\bf Output:} 
updated parameters $\theta_Q$, $\theta_{\pi^i}$, $i = 1,\ \cdots,\ N_p$
\begin{algorithmic}[1]
	\For{i=1,$\cdots$, $N_p$} 
	\State Sample a random mini-batch of $N$ transitions $(x_k, u_k, r_k, x_{k+1})$ from $R^i$
	\State Set $y_k=r_k+\gamma \bar{Q}(x_{k+1},\bar{\pi}^i(x_{k+1}|\theta_{\bar{\pi}}^i)|\theta_{\bar{Q}})$
	\State Update the shared critic by minimizing the loss: 
	\begin{equation*}
	L _Q=  \mathbb{E}_{x_k}[(y_k-Q(x_k,u_k|\theta_Q))^2]
	\end{equation*}
	\State Update the actor policy by minimizing the loss: 
		\begin{equation*}
	L _{\pi^i}=  \mathbb{E}_{x_k}[Q(x_k,u_k|\theta_Q)|_{u_k=\pi^i(x_k)}]+\lambda KL[\pi^i_{old}|\pi^i]
	\end{equation*}
\EndFor
\end{algorithmic}
\end{algorithm}

However, it is not possible to straightforwardly apply DDPG to our neural network structure, because, as mentioned before, the PLCBC architecture has multiple policy subnetworks (the location network is frozen during reinforcement learning). 
The multiple policy subnetworks will perform as multiple actor in the framework of DDPG.
When using this special network for reinforcement learning, the challenge is that action executed by agent in each time step may be generated by different policy subnetwork. 
It is not reasonable that all the policy subnetworks update its weights based on the gradient generated by experience sample from other policy. 
Our solution is quite simple and intuitive. 
Instead of single replay buffer, we use multiple replay buffer $R^i​$, one for each policy subnetwork. 
The tuple $ (x_t, u_t, r_t, x_{t+1}) ​$  is stored in the replay buffer $R^i​$ if and only if $x_t\in P^i​$, and the gradient estimated by this transition will only applied in the corresponding policy subnetwork. 
Besides that, in order to take fully advantage of the sampled data, we use a shared critic network to estimate the value of each state-action pair. 
This means that, in each iteration, the single critic network will update its weight by using the transition from different policy with Q-learning method.
The modified algorithm is described in Algorithm 1,2.

\section{Experimental Evaluation}

The primary focus of our experiment evaluation is to demonstrate that our algorithm is safe and sample efficient for a diverse range of control problems.
We compared the efficiency of our method against several prior methods with continuous state and action spaces.
Besides using accurate system model, we also have verified that PLCBC can still have better performance even with a certain degree of  model mismatch.

\subsection{Experimental Domains and Setup}

In order to verify the efficiency of PLCBC, we compare it to several prior methods ,including MPC, Distributed Proximal Policy Optimization (DPPO) \cite{heess2017emergence}, DDPG, and DDPG pre-trained by supervised learning (SP+DDPG). 

The comparisons are conducted on three test environments, namely pendulum swing-up, quadcopter navigation and urban traffic network control.
First, we evaluate our algorithm on inverted pendulum from OpenAI Gym \cite{openaigym}.
The goal of this task  is  to remain the pendulum at zero angle (vertical), with the least rotational velocity, and the least effort.
Quadcopter  navigation domain is to control the quadcopter to track a fixed trajectory. 
The dynamic system used in this domain is a quadcopter model described in \cite{bemporad2009}.
The final experimental domain is urban traffic network control problem which aims to alleviate congestion by adjusting green time ratio.
We use traffic network modeler Paramics to simulate a small 5-junctions unban traffic system. 

The time horizon $N$ of MPC is 3 time steps.
Other reinforcement learning algorithms (DPPO, DDPG, SP+DDPG) use non-linear fully connected neural networks and have similar number  of total parameters to PLCBC.
SP+DDPG use supervised learning to pre-train the policy.
In this case, we use MPC controller to generate training data.
Supervised pre-training is expected to offer a good starting point which can then be fine-tuned using DDPG \cite{supervised}.

\subsection{Empirical Results}
\begin{figure}[htp]
\begin{center}
	\subfigure[Pendulum]{\includegraphics[width=0.47\textwidth]{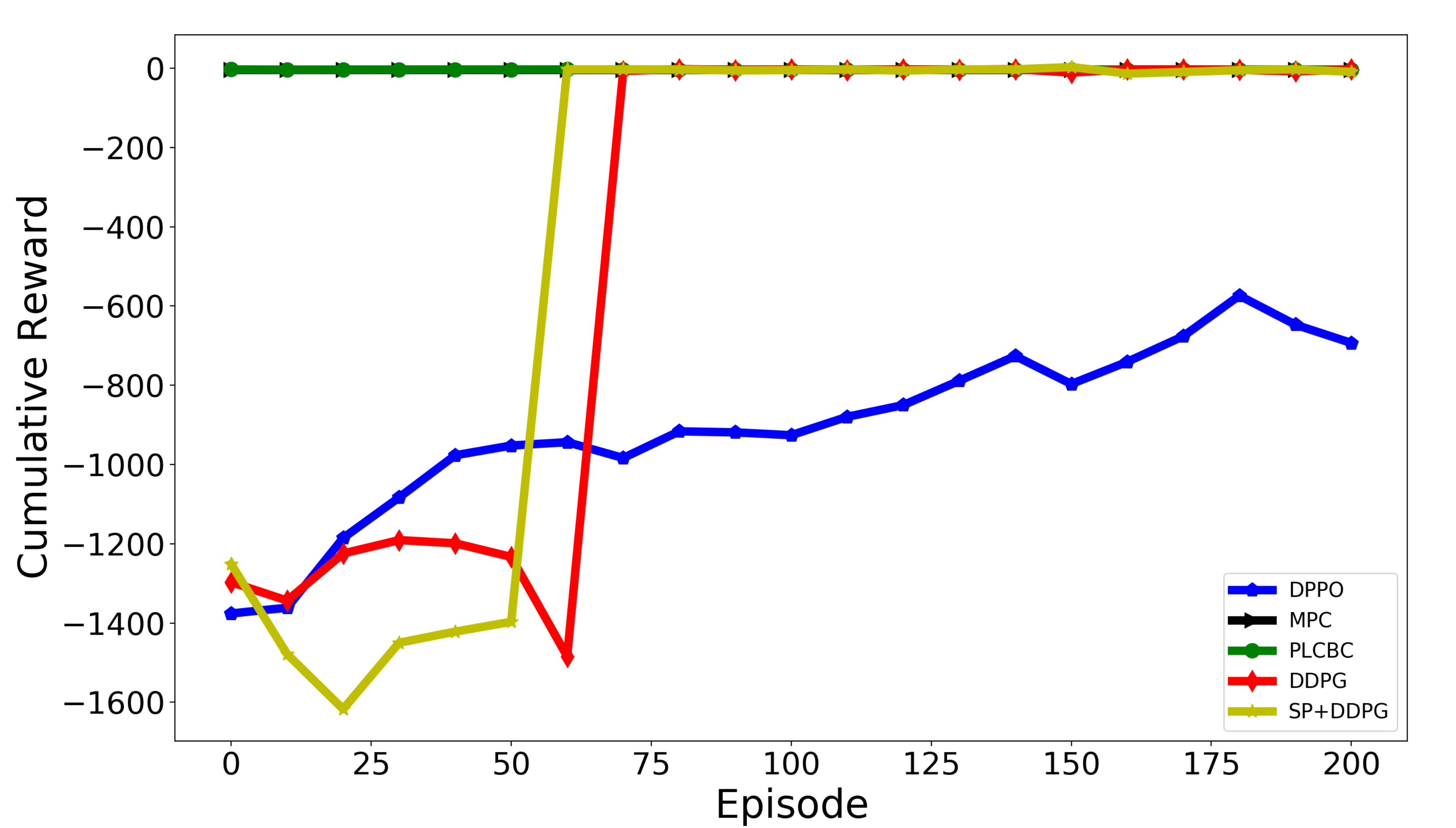}}
	\subfigure[Quadcopter Navigation]{\includegraphics[width=0.47\textwidth]{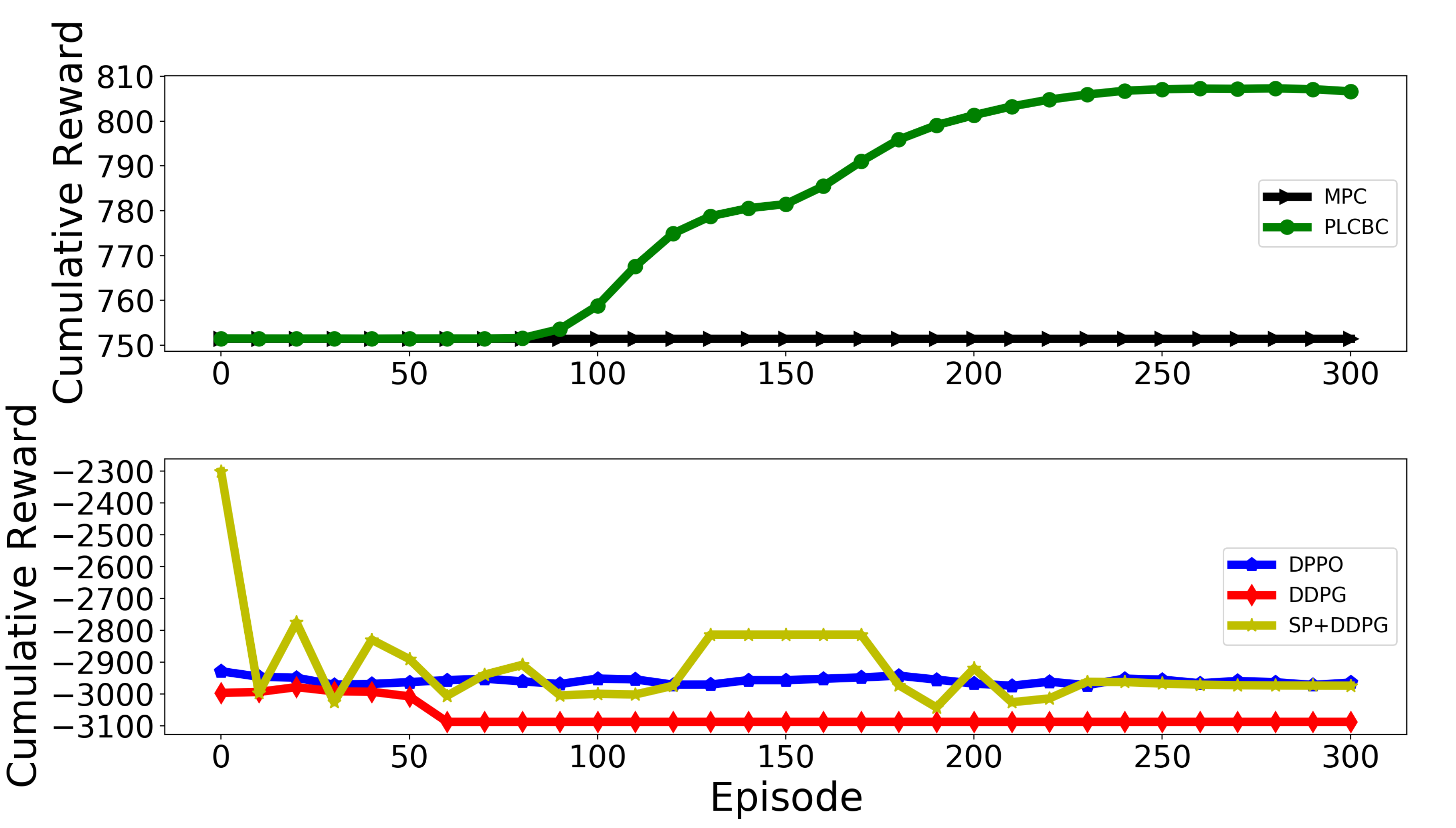}}
	\caption{Performance (Cumulative Reward on y-axis) versus number of episodes (Episode on x-axis) of DDPG, MPC (without model error), SP+DDPG, PLCBC (without model error), DPPO on different domains}\label{fig:perfect_results}
\label{fig:withouterror}
\end{center}
\end{figure}
We set the max number of training episodes $M = 200$ , the max number of steps of each episode $T = 200$ for  pendulum domain, and $M = 300$, $T = 200$ for quadcopter navigation and traffic network control domain.
The performance criteria is the accumulation reward within the max number of steps $T$.
Fig.~\ref{fig:withouterror} shows the cumulative reward after execution of each algorithm over $M$ iterations.
As it can be seen from the Fig.~\ref{fig:withouterror}, in both domains, PLCBC can perform really well without any training, and slightly improve its performance along the iterations.
DDPG, on te other hand, has to learn from a lot of mistakes and needs much more iterations to converge.
Besides that, as shown in Fig.~\ref{fig:withouterror}(b),  DDPG is prone to converge at  a worse suboptimal solution and  can not improve its performance at all.
As a widely used trick, SP+DDPG uses the MPC's experience samples to pre-train random policy.
However, in both domains, SP+DDPG can not extremely improve the training efficiency of  DDPG.
The reason is that  this kind of supervised learning based method can not generalize to all states encountered during training, especially in continuous action space.
\begin{figure}[htp]
\begin{center}
	\includegraphics[width=0.6\textwidth]{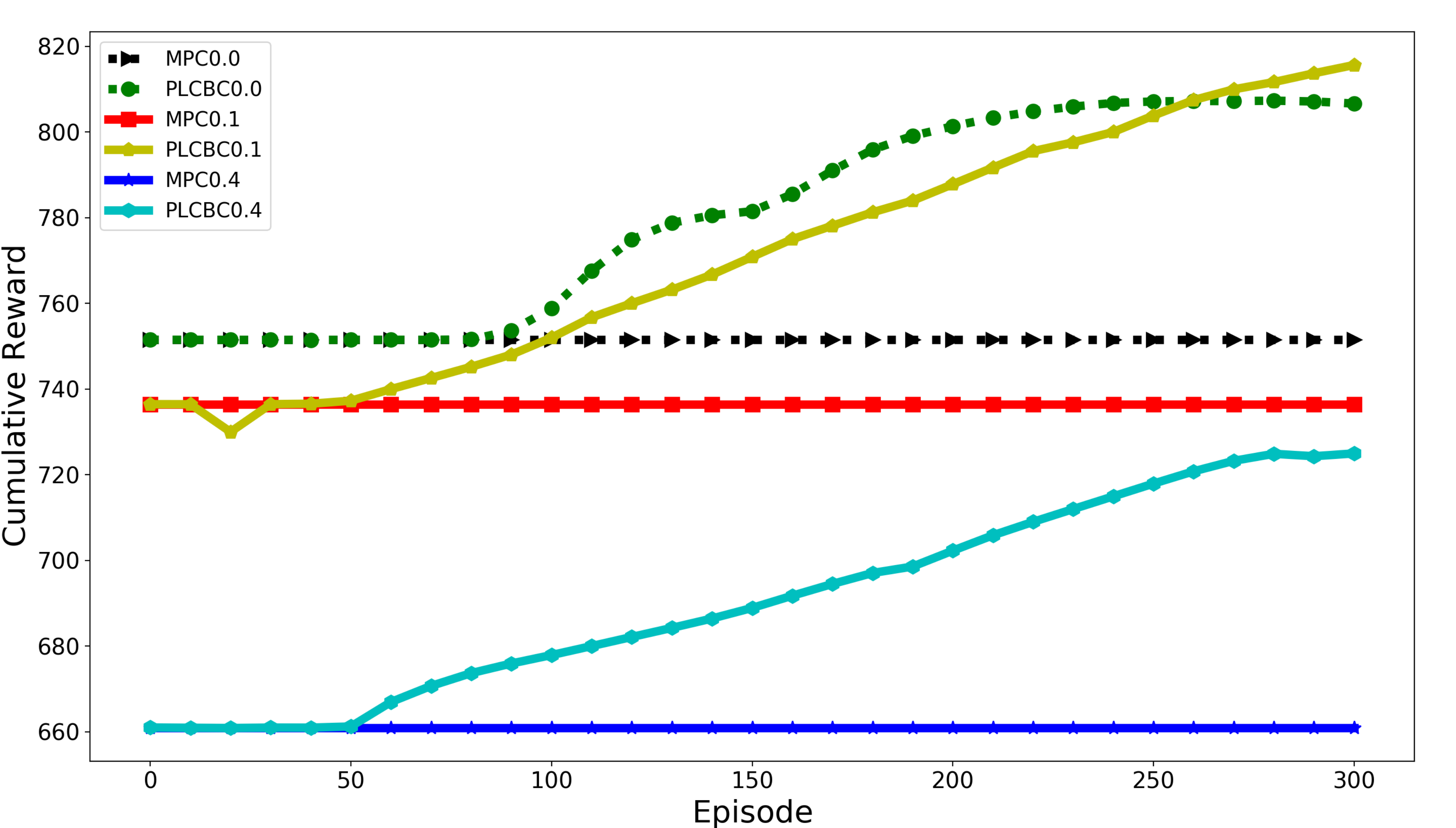}
	\caption{Performance (Cumulative Reward on y-axis) versus number of episodes (Episode on x-axis) of DDPG, MPC (with model error), SP+DDPG, PLCBC (with model error) on quadcopter domain. MPC$\{0.0,  0.1, 0.4\}$ and PLCBC$\{0.0,  0.1, 0.4\}$ indicates algorithm with mass error $\epsilon = 0.0, 0.1, 0.4$.}\label{fig:witherror_results}
\label{fig:witherror}
\end{center}
\end{figure}

In the previous set of experiment, it is assumed that an accurate model is available.
There is another experiment to show how the new method can yield robust feedback behavior even in the presence of model error.
To  illustrate that, we modified the quadcopter model by decreasing the mass of the quadcopter by $10\%$ and $40\%$ denoted by $\epsilon = 0.1, 0.4$.
We only provide comparisons to MPC baseline (with various model error), since DDPG, SP+DDPG are model free, and will have the same result as last set of experiment. 
Rich theory has been developed to verify that stability is maintained for a specified range of model variations and a class of noise signals  in the context of MPC \cite{afram2014theory}.
Therefore,  as shown in Fig.~\ref{fig:witherror}, MPC is robust to slightly model error.
However, it can not recover from mistakes when the model error is relatively large.
As for PLCBC, although the initial performance gets worse in the presence of model error, it still can improve its performance along with the iterations within a specified range of model error.

\begin{figure}[htp]
\begin{center}
	\subfigure[]{\includegraphics[width=0.4\textwidth]{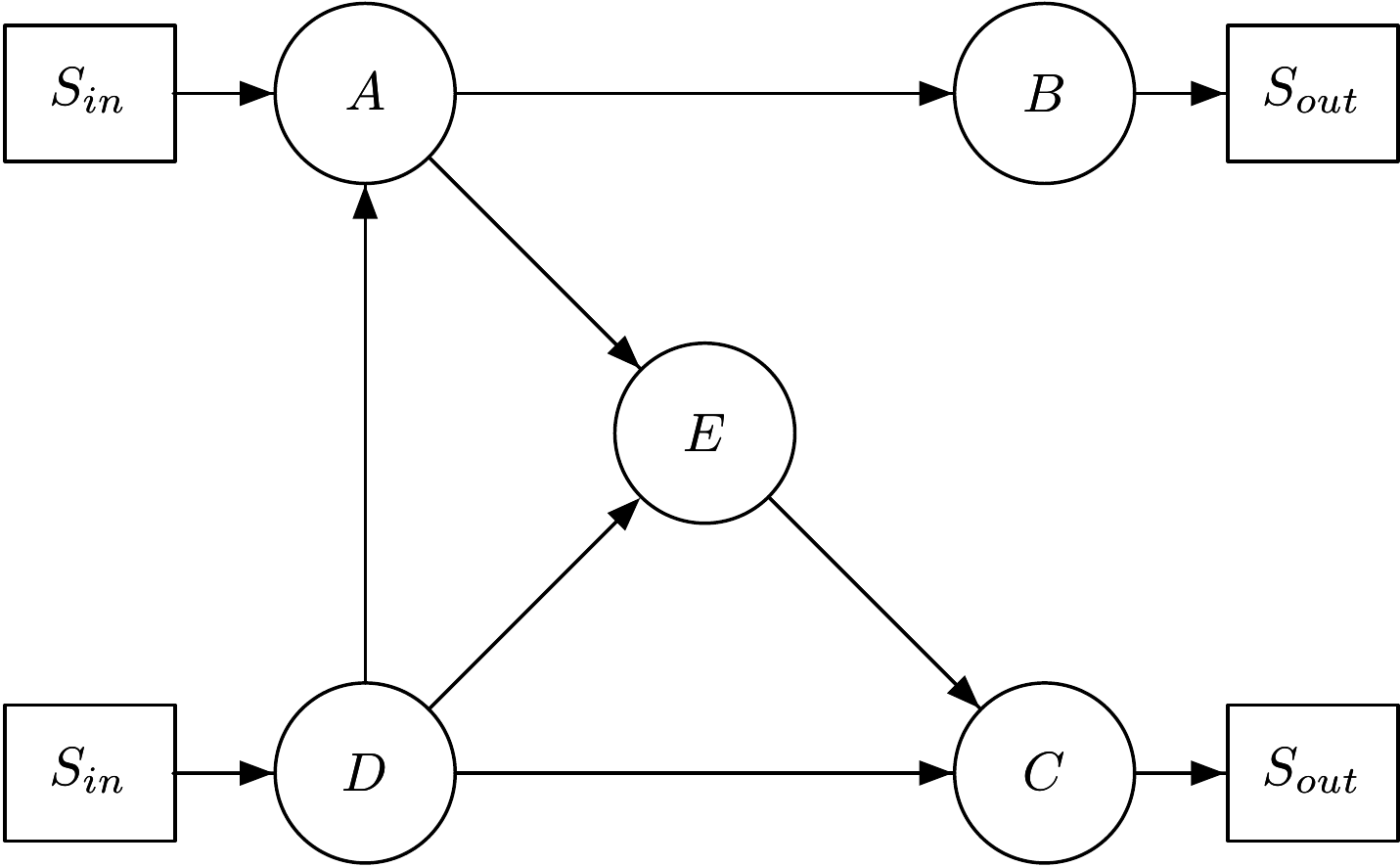}}
	\subfigure[]{\includegraphics[width=0.47\textwidth]{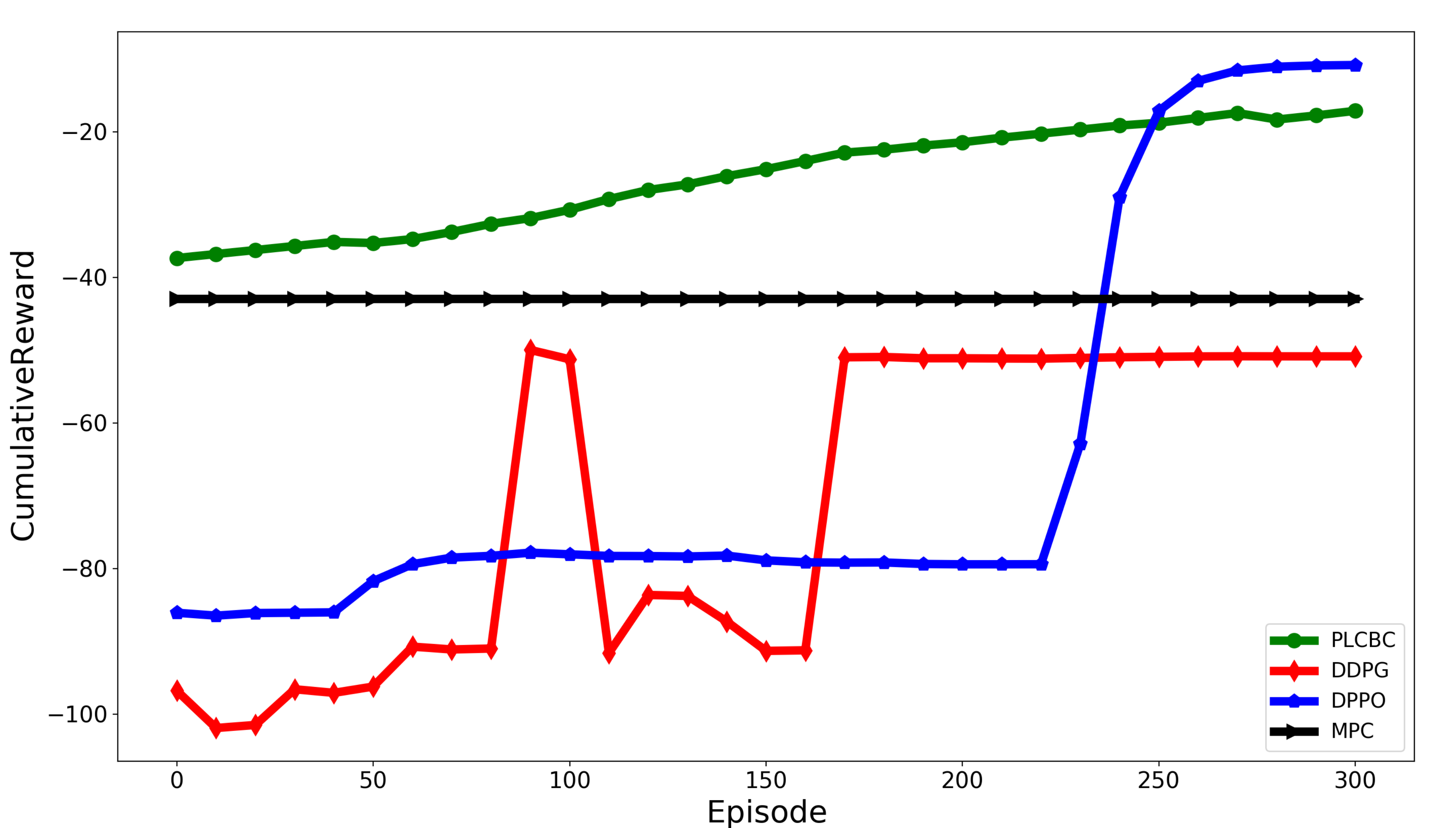}}
	\caption{(a) A toy traffic network for simulation experiments. (b) Performance (Cumulative Reward on y-axis) versus number of episodes (Episode on x-axis) of DDPG, MPC, PLCBC, DPPO on urban traffic network control domain.}
\label{fig:trafictrain}
\end{center}
\end{figure}

\begin{figure}[htp]
\begin{center}
	\subfigure[MPC]{\includegraphics[width=0.48\textwidth]{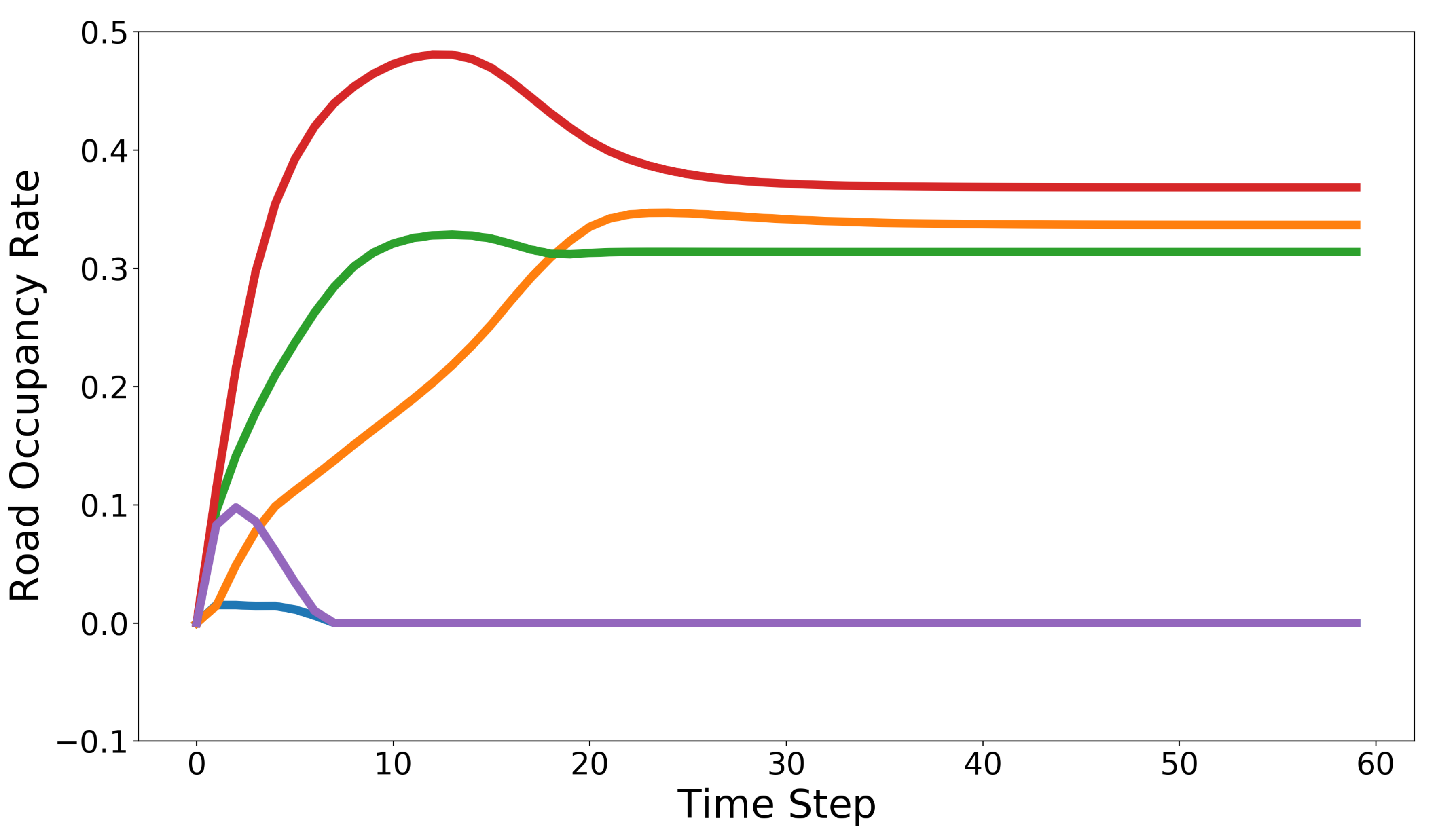}}
	\subfigure[PLCBC]{\includegraphics[width=0.48\textwidth]{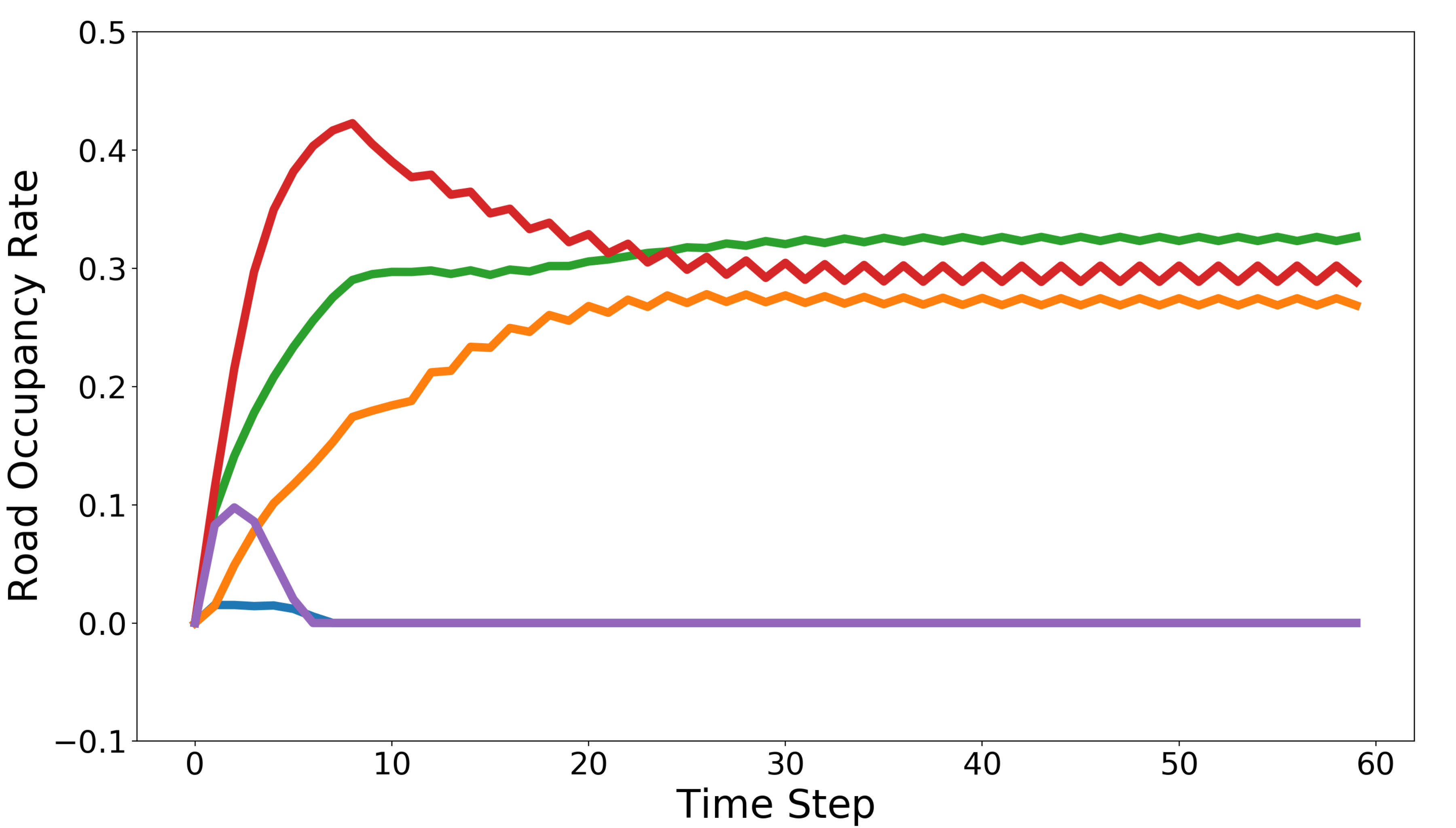}}
	\caption{Performance (Road Occupancy Rate of each lane on y-axis) versus number of time steps (Time Step on x-axis) of MPC and PLCBC.}
\label{fig:roadocu}
\end{center}
\end{figure}

As has been seen, PLCBC is robust to model mismatch.
It makes this method really appropriate to urban traffic system control problem which suffers from external disturbances and highly nonlinearity of dynamic system.
In this set of experiment, we verify the efficiency of PLCBC in alleviating traffic congestion.
Our toy traffic network model, shown as in Fig.~\ref{fig:trafictrain}(a), was created by the traffic network modeler, Paramics.
This traffic network consists of 5 signalized junctions $A, B, C, D, E$.
The node $S_{in}$ and $S_{out}$ are defined as the source node and sink node respectively.
The comparison results during training phase are shown in Fig.~\ref{fig:trafictrain} (b).
As shown in plot, PLCBC has a great initial solution, and can improve its performance by learning from interaction experience which is safe to perform in the real-world system.
However, DDPG and DPPO has to learn from scratch.
After the training phase, MPC and fully trained policy PLCBC are deployed to the simulative traffic network.
As shown in Fig.~\ref{fig:roadocu}, MPC controller tends to lead vehicles into some specific links, while PLCBC has learned to balance the traffic burden.  

\section{DISCUSSION AND FUTURE WORK}
In this paper, a sample efficient reinforcement learning algorithm is proposed to initialize the neural network with an optimal MPC controller.
Unlike the method of supervised learning, PLCBC directly transforms the MPC controller into neural network by using multi-parametric technique.
Our empirical evaluation has verified that PLCBC greatly improves the  convergence properties of traditional methods, and can handle some complex control problem, such as quadcopter navigation.
However, PLCBC still has some deficiencies worthy of further improvement.
First, since PLCBC is a shallow neural network with only 2 layers, a large number of neurons are required to express all regions.
It has been proved that multi-layer neural network can form more regions in the state space with the same number of neurons \cite{montufar2014number}.
Therefore, a meaningful research direction is how to use multi-layer neural network to express the region structure.
In addition, traditional reinforcement learning algorithm tends to learn from random actions, and worsen the effectiveness of policy initialized by a optimal controller like MPC.
Therefore, how to design an efficient reinforcement learning algorithm to incorporate with PLCBC is also a promising issue.
\section*{REFERENCE}
\bibliography{mybib}{}  

\begin{thebibliography}{10}

\bibitem{Silver:2016aa}
D.~Silver, A.~Huang, C.J. Maddison, A.~Guez, and L.~Sifre et~al.
\newblock Mastering the game of go with deep neural networks and tree search.
\newblock {\em Nature}, 529(7587):484--489, 2016.

\bibitem{Silver:2017aa}
D.~Silver, J.~Schrittwieser, K.~Simonyan, I.~Antonoglou, and A.~Huang et~al.
\newblock Mastering the game of go without human knowledge.
\newblock {\em Nature}, 550(7676):354--359, 2017.

\bibitem{levine2016end}
S.~Levine, C.~Finn, T.~Darrell, and P.~Abbeel.
\newblock End-to-end training of deep visuomotor policies.
\newblock {\em The Journal of Machine Learning Research}, 17(1):1334--1373,
  2015.

\bibitem{schulman2017proximal}
S.~John, W.~Filip, D.~Prafulla, R.~Alec, and K.~Oleg.
\newblock Proximal policy optimization algorithms.
\newblock {\em arXiv preprint arXiv:1707.06347}, 2017.

\bibitem{heess2017emergence}
N.~Heess, T.B. Dhruva, S.~Sriram, J.~Lemmon, J.~Merel, G.~Wayne, Y.~Tassa,
  T.~Erez, Z.~Wang, and S.M.A. Eslami.
\newblock Emergence of locomotion behaviours in rich environments.
\newblock {\em arXiv preprint arXiv:1707.02286}, 2017.

\bibitem{gu2016q}
S.~Gu, T.~Lillicrap, Z.~Ghahramani, R.E. Turner, and S.~Levine.
\newblock Q-prop: Sample-efficient policy gradient with an off-policy critic.
\newblock {\em Conference on Learning Representations (ICLR)}, 2017.

\bibitem{levine2013guided}
S.~Levine and V.~Koltun.
\newblock Guided policy search.
\newblock {\em Proceedings of the 30th International Conference on Machine
  Learning}, 28(3):1--9, 2013.

\bibitem{XI2013}
Y.G. XI, D.W. LI, and S.~LIN.
\newblock Model predictive control — status and challenges.
\newblock {\em Acta Automatica Sinica}, 39(3):222 -- 236, 2013.

\bibitem{Schroecker2017}
Y.~Schroecker and C.L. Isbell.
\newblock State aware imitation learning.
\newblock {\em Advances in Neural Information Processing Systems(NIPS)}, 2017.

\bibitem{Wang2017}
Z.~Wang, J.S. Merel, S.E. Reed, N.~de~Freitas, G.~Wayne, and N.~Heess.
\newblock Robust imitation of diverse behaviors.
\newblock {\em Advances in Neural Information Processing Systems(NIPS)}, 2017.

\bibitem{Mnih2013}
V.~Mnih, K.~Kavukcuoglu, D.~Silver, A.~Graves, I.~Antonoglou, D.~Wierstra, and
  M.~Riedmiller.
\newblock Playing atari with deep reinforcement learning.
\newblock {\em NIPS Deep Learning Workshop}, 2013.

\bibitem{DDPG}
T.P. Lillicrap, J.J. Hunt, A.~Pritzel, N.~Heess, T.~Erez, Y.~Tassa, D.~Silver,
  and D.~Wierstra.
\newblock Continuous control with deep reinforcement learning.
\newblock {\em 6th International Conference on Learning Representations}, 2016.

\bibitem{Dagger2011}
S.R. Ross, G.J. Gordon, and D.~Bagnell.
\newblock A reduction of imitation learning and structured prediction to
  no-regret online learning.
\newblock {\em Proceedings of the fourteenth international conference on
  artificial intelligence and statistics}, pages 627--635, 2011.

\bibitem{Levine}
S.~Levine and P.~Abbeel.
\newblock Learning neural network policies with guided policy search under
  unknown dynamics.
\newblock {\em Advances in Neural Information Processing Systems}, pages
  1071--1079, 2014.

\bibitem{fazel2018global}
X.~Chen and S.~Cai.
\newblock Global convergence of policy gradient methods for linearized control
  problems.
\newblock {\em 2016 35th Chinese Control Conference}, 2018.

\bibitem{Florensa2017}
C.~Florensa, D.~Held, M.~Wulfmeier, M.~Zhang, and P.~Abbeel.
\newblock Reverse curriculum generation for reinforcement learning.
\newblock {\em arXiv preprint arXiv:1707.05300}, pages 1--14, 2017.

\bibitem{ross2010efficient}
S.~Ross and D.~Bagnell.
\newblock Efficient reductions for imitation learning.
\newblock {\em Proceedings of the thirteenth international conference on
  artificial intelligence and statistics}, pages 661--668, 2010.

\bibitem{Kahn}
G.~Kahn, T.~Zhang, S.~Levine, and P.~Abbeel.
\newblock Plato: Policy learning using adaptive trajectory optimization.
\newblock pages 3342--3349, 2017.

\bibitem{Argall2009}
B.D. Argall, S.~Chernova, M.~Veloso, and B.~Browning.
\newblock {A survey of robot learning from demonstration}.
\newblock {\em Robotics and Autonomous Systems}, 57(5):469--483, 2009.

\bibitem{guo2014}
X.~Guo, S.~Singh, H.~Lee, R.L. Lewis, and X.~Wang.
\newblock Deep learning for real-time atari game play using offline monte-carlo
  tree search planning.
\newblock pages 3338--3346, 2014.

\bibitem{Bagnell2010}
R.~Stephane and B.~Drew.
\newblock Efficient reductions for imitation learning.
\newblock {\em Proceedings of the 13th International Conference on Artificial
  Intelligence and Statistics}, 9:661--668, 2010.

\bibitem{Zhang}
T.~Zhang, G.~Kahn, S.~Levine, and P.~Abbeel.
\newblock Learning deep control policies for autonomous aerial vehicles with
  mpc-guided policy search.
\newblock {\em Proceedings of IEEE International Conference on Robotics and
  Automation}, 2015.

\bibitem{levine2014b}
S.~Levine and V.~Koltun.
\newblock Learning complex neural network policies with trajectory
  optimization.
\newblock {\em International Conference on Machine Learning}, pages 829--837,
  2014.

\bibitem{levine2015b}
S.~Levine, N.~Wagener, and P.~Abbeel.
\newblock Learning contact-rich manipulation skills with guided policy search.
\newblock {\em IEEE International Conference on Robotics and Automation}, pages
  156--163, 2015.

\bibitem{Bemporad2002}
A.~Bemporad, M.~Morari, V.~Dua, and E.N. Pistikopoulos.
\newblock The explicit linear quadratic regulator for constrained systems.
\newblock {\em Automatica}, 38(1):3--20, 2002.

\bibitem{1184260}
T.A. Johansen.
\newblock On multi-parametric nonlinear programming and explicit nonlinear
  model predictive control.
\newblock {\em Proceedings of the 41st IEEE Conference on Decision and
  Control}, 3:2768--2773, 2002.

\bibitem{GP2007}
G.~Pannocchia, J.B. Rawlings, and S.J. Wright.
\newblock Brief paper: Fast, large-scale model predictive control by partial
  enumeration.
\newblock {\em Automatica}, 43(5):852--860, 2007.

\bibitem{openaigym}
G.~Brockman, V.~Cheung, L.~Pettersson, J.~Schneider, J.~Schulman, J.~Tang, and
  W.~Zaremba.
\newblock Openai gym.
\newblock {\em arXiv preprint arXiv：1606.01540}, 2016.

\bibitem{bemporad2009}
A.~Bemporad, C.A. Pascucci, and C.~Rocchi.
\newblock Hierarchical and hybrid model predictive control of quadcopter air
  vehicles.
\newblock {\em IFAC Proceedings Volumes}, 42(17):14--19, 2009.

\bibitem{supervised}
P.H. Su, P.~Budzianowski, S.~Ultes, M.~Gasic, and S.~Young.
\newblock Sample-efficient actor-critic reinforcement learning with supervised
  data for dialogue management.
\newblock {\em Proceedings of the 18th Annual SIGdial Meeting on Discourse and
  Dialogue}, pages 147--157, 2008.

\bibitem{afram2014theory}
A.~Afram and F.~Janabi-Sharifi.
\newblock Theory and applications of hvac control systems--a review of model
  predictive control (mpc).
\newblock {\em Building and Environment}, 72:343--355, 2014.

\bibitem{montufar2014number}
G.F. Montufar, R.~Pascanu, K.~Cho, and Y.~Bengio.
\newblock On the number of linear regions of deep neural networks.
\newblock pages 2924--2932, 2014.

\end{thebibliography}
\bibliographystyle{unsrt}

\end{document}